\documentclass[10pt,twocolumn,letterpaper]{article}

\usepackage{iccv}
\usepackage{times}
\usepackage{epsfig}
\usepackage{graphicx}
\usepackage{amsmath}
\usepackage{amssymb}
\usepackage{adjustbox}
\usepackage[normalem]{ulem}

\usepackage{caption} % for captionof{} of wide figure in the first page
\usepackage{booktabs} % for table rules
\usepackage[ruled,vlined]{algorithm2e} % for algorithms
\usepackage{lipsum} % for text placeholder
\usepackage{multirow, makecell}
\usepackage{enumitem}
\usepackage{graphicx}
\usepackage{animate}
\usepackage[table, usenames, dvipsnames]{xcolor}

\usepackage[pagebackref=true,breaklinks=true,colorlinks,bookmarks=false]{hyperref}

\iccvfinalcopy % *** Uncomment this line for the final submission

\def\iccvPaperID{****} % *** Enter the ICCV Paper ID here

\ificcvfinal\fi

\newcommand{\OURS}{Sat2Vid}
\newcommand{\USEADOBE}{Please use \textbf{Adobe Reader} / \textbf{KDE Okular} to see animations.}
\newcommand{\NUMFRAMEA}{14} % 14
\newcommand{\NUMFRAMEB}{59} % 59

\newcommand{\todo}[1]{\textcolor{black}{#1}}
\newcommand{\boldparagraph}[1]{\vspace{0.1em}\noindent{\bf #1} }

\newcolumntype{L}[1]{>{\raggedright\let\newline\\\arraybackslash\hspace{0pt}}m{#1}}
\newcolumntype{C}[1]{>{\centering\let\newline\\\arraybackslash\hspace{0pt}}m{#1}}
\newcolumntype{R}[1]{>{\raggedleft\let\newline\\\arraybackslash\hspace{0pt}}m{#1}}

\begin{document}

%%%%%%%%% TITLE
\title{Sat2Vid: Street-view Panoramic Video Synthesis from a Single Satellite Image}

\author{
Zuoyue Li$^\text{1}$, Zhenqiang Li$^\text{2}$, Zhaopeng Cui$^\text{3}$, Rongjun Qin$^\text{4}$, Marc Pollefeys$^\text{1,5}$, Martin R. Oswald$^\text{1}$\\
$^\text{1}$ETH Zurich \hspace{1em} $^\text{2}$The University of Tokyo \hspace{1em} $^\text{3}$Zhejiang University\\
$^\text{4}$The Ohio State University \hspace{1em} $^\text{5}$Microsoft
}

\maketitle
\ificcvfinal\thispagestyle{empty}\fi

%%%%%%%%% ABSTRACT
\begin{abstract}
We present a novel method for synthesizing both temporally and geometrically consistent street-view panoramic video from a single satellite image and camera trajectory. Existing cross-view synthesis approaches focus on images, while video synthesis in such a case has not yet received enough attention. For geometrical and temporal consistency, our approach explicitly creates a 3D point cloud representation of the scene and maintains dense 3D-2D correspondences across frames that reflect the geometric scene configuration inferred from the satellite view.
\todo{As for synthesis in the 3D space, we implement a cascaded network architecture with two hourglass modules to generate point-wise coarse and fine features from semantics and per-class latent vectors, followed by projection to frames and an upsampling module to obtain the final realistic video.}
%For realistic synthesis, we implement a cascaded network architecture with two hourglass modules to generate colored frames from semantics and per-class latent vectors that are projected from 3D point clouds.
By leveraging computed correspondences, the produced street-view video frames adhere to the 3D geometric scene structure and maintain temporal consistency. Qualitative and quantitative experiments demonstrate superior results compared to other state-of-the-art synthesis approaches that \todo{either lack temporal consistency or realistic appearance}. To the best of our knowledge, our work is the first one to synthesize cross-view images to video.
\end{abstract}
\vspace{-1.5em}

%To this end, our approach explicitly creates a 3D point cloud representation of the scene and maintains dense 3D-2D correspondences across frames that reflect the geometric scene configuration inferred from the satellite view.

\section{Introduction}

Street-view images have been proven to be helpful for exploring remote places or for strategic ground planning in emergency or intelligence operations. 
They are useful for a variety of applications in virtual or mixed reality, realistic simulations and gaming, viewpoint interpolation, or cross-view matching. 
Nevertheless, their acquisition is rather expensive, and regular updates to capture changes are required for some tasks.
On the other hand, satellite images are regularly captured, easier to obtain, have significantly better earth coverage, and are generally much more widely available than street-view images.
The automatic generation of street-view images from given satellite or aerial images is thus an attractive and interesting alternative for several aforementioned applications.

While single street-view image generation from satellite images has recently been investigated~\cite{regmi2018cross,lu20geometry}, these methods are not suitable to create continuous view-point changes around a given location since they built upon random generators and lack constraints on the correspondence between frame pixels.
They are thus unable to synthesize temporally and geometrically consistent image sequences that are desired for a better visual experience.

\begin{figure}[t]
    \centering
    \footnotesize
    \newcommand{\sz}{0.33} % figure size
    % \noindent\animategraphics[autoplay,loop,height=\sz\linewidth]{5}{fig/forward/slt_sate_frames/86/}{00}{\NUMFRAMEA}
    % \noindent\animategraphics[autoplay,loop,height=\sz\linewidth]{5}{fig/forward/slt_ours_jpg/86/}{00}{\NUMFRAMEA}\\ [1pt]
    \noindent\animategraphics[autoplay,loop,height=\sz\linewidth]{5}{fig/forward/slt_sate_frames/178/}{00}{\NUMFRAMEA}
    \noindent\animategraphics[autoplay,loop,height=\sz\linewidth]{5}{fig/forward/slt_ours_jpg/178/}{00}{\NUMFRAMEA}\\ [1pt]
    \noindent\animategraphics[autoplay,loop,height=\sz\linewidth]{5}{fig/forward/slt_sate_frames/284/}{00}{\NUMFRAMEA}
    \noindent\animategraphics[autoplay,loop,height=\sz\linewidth]{5}{fig/forward/slt_ours_jpg/284/}{00}{\NUMFRAMEA}\\[-3pt]
    \setlength{\tabcolsep}{1pt}
    \begin{tabular}{C{77pt}C{154pt}}
      Input Satellite RGB & Synthesized Video \\[-10pt]
    \end{tabular}
    \caption{\textbf{Street-view panoramic video synthesis results of our method}. For a single satellite image and a given trajectory (indicated by a red arrow in the figure), we learn to synthesize a corresponding street-view panoramic video with both geometrical and temporal consistency. \USEADOBE}
    \label{fig:teaser}
    \vspace{-1em}
\end{figure}

In this paper, we approach the novel task to synthesize street-view panoramic video sequences as realistically as possible and as consistent as possible from a single satellite image and given viewing locations. 
To achieve this, instead of resorting to 2D generators like \cite{regmi2018cross,lu20geometry} and generating images individually, we propose to generate the whole scene in a 3D representation of point cloud, and establish the correspondence between these visible points and the 2D frame pixels. In this way, the projected views from the entire generated scene instance will be naturally consistent by design.
In order to generate image frames as good as a single image, we design a two-stage 3D generator in a coarse-to-fine manner that exploits the characteristics of different 3D convolutional neural networks.
Fig.~\ref{fig:teaser} presents two examples of our synthesized results, which well demonstrate the temporal consistency of our generated video.

Our major \textbf{contributions} can be summarized as follows. 
\textbf{(1)} We present the first work for satellite-to-ground video synthesis from a single satellite image with a trajectory. 
\textbf{(2)} We propose a novel cross-view video synthesis method that ensures both spatial and temporal consistency by explicitly modeling a cross-frame correspondence using a 3D point cloud representation and building projective geometry constraints into our network architecture. 
\textbf{(3)} Our method outperforms multiple baseline methods both qualitatively and quantitatively on a newly-constructed dataset for cross-view video synthesis that is expanded from the London panorama dataset~\cite{lu20geometry}.
The source code and pre-trained models will be made publicly available upon publication.
\section{Related Work}

\begin{figure*}
    \vspace{-3pt}
	\centering
	\includegraphics[width=\linewidth]{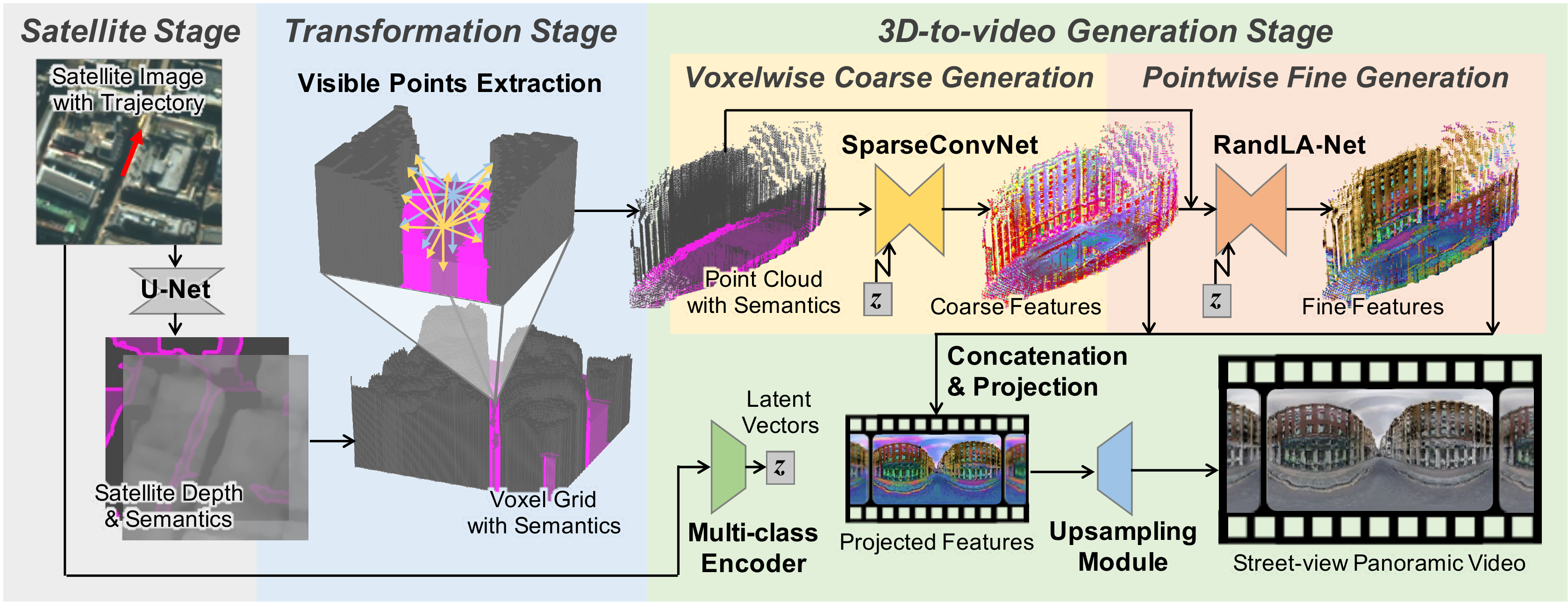}
	\vspace{-17pt}
	\caption{
	\textbf{Overview of our network architecture.} Our network consists of multiple sub-networks accounting for three processing stages which transform between different scene representations. These stages are:
	\textbf{Satellite Stage:} The input satellite image is processed by a 2D U-Net~\cite{ronneberger15unet} to generate a 2.5D height map with corresponding semantics.
	\textbf{Transformation Stage:} To obtain a 3D representation, the semantic height map is converted into a semantic voxel occupancy grid. Visible points are then extracted according to the sampling points of the input trajectory.
	\textbf{3D-to-video Generation Stage:} A generator operating in the 3D domain infers features for each point from the semantics.
	The cascaded SparseConvNet~\cite{graham18sparse} and RandLA-Net~\cite{hu20randlanet} both with hourglass structures act on coarse and fine generation successively.
	Rather than using a single seed as in \cite{lu20geometry}, we use a multi-class texture encoder that computes multiple latent vectors from the input satellite image.
	Lastly, the point cloud with concatenated features is projected to each frame, which is finally upsampled with a light-weight network to double the resolution.
    \textbf{Note: (1)} The 3D-to-video generation stage is trained under the framework of BicycleGAN~\cite{zhu2017toward};
    \textbf{(2)} Sky points are included in the pipeline but not visualized here; 
    \textbf{(3)} Features are illustrated with pseudo colors.
	}
	\vspace{-1em}
	\label{fig:pipeline}
\end{figure*}
\textbf{Cross-view synthesis} focuses on synthesizing from a completely different view of the given image. Most existing works in this field are targeted at single image synthesis. A very typical application is to generate the street view from a given satellite image. Zhai~\etal~\cite{zhai2017predicting} proposed to learn to map the semantic segmentation from the aerial to the ground perspective, which can be further used to synthesize ground-level views based on GANs~\cite{goodfellow2014generative}. Regmi~\etal~\cite{regmi2018cross, regmi2019bridging} proposed to use conditional GANs to learn the aerial or ground view images together with semantic segmentation. In order to keep the geometrical consistency, Lu~\etal~\cite{lu20geometry} proposes a differentiable geo-transformation layer that turns a semantically labeled satellite depth image to corresponding street-view depth and semantics for further street-view panorama generation. Turning to the field of cross-view video synthesis, there is no much work involving in yet as the problem becomes even harder. Although the video can be synthesized frame-by-frame by the image synthesis method, its temporal consistency is hard to be guaranteed, which is important for a video.

\textbf{Video synthesis} is a field that attracted more attentions in the community and have various forms according to the given input, which can be roughly divided into the following three categories. 
\textbf{(1)} Unconditional video synthesis \cite{logacheva2020deeplandscape,saito2017temporal,tulyakov2018mocogan,vondrick2016generating} generates video clips from given input random variables by extending the current GAN frameworks on (spatial) images further into the temporal dimension. 
\textbf{(2)} Future video prediction \cite{finn2016unsupervised,hao2018controllable,li2018flow,liang2017dual,lotter2016deep,mathieu2015deep,pan2019video,walker2016uncertain,walker2017pose} aims at inferring the future frames of a video based on the current observations so far. 
\textbf{(3)} Video-to-video synthesis~\cite{chan2019everybody,chen2019mocycle,mallya2020world,wang2019few,wang2018vid2vid} is closer to our task, which
%\sout{maps a video in a certain domain to what in another domain, which shares a similar spirit with image-to-image translation with enhanced temporal consistency.} 
maps a video from a source domain to a target domain (\eg, generating RGB images from a sequence of semantic segmentation masks or depth images). 
Compared to the image-to-image translation task, it emphasizes the coherency of the generated video frames over time. 
Wang~\etal~\cite{wang2018vid2vid} aimed to achieve this by leveraging a generative adversarial learning framework and spatio-temporal adversarial objective. 
Mallya~\etal~\cite{mallya2020world} proposed an enhanced method that achieves consistency over a longer time by a guidance image projected from an incrementally colored point cloud during the subsequent frame generation. 
Nevertheless, the cross-view video synthesis setting in our work is still different from all these categories, which should consider both the temporal consistency between video frames and the geometrical consistency between top and ground views.

\textbf{Novel view synthesis and neural rendering} technologies develop rapidly recently with the advancements in deep neural networks. 
Many state-of-the-art works focus on the synthesis from a single image. SynSin~\cite{wiles2020synsin} proposed an end-to-end view synthesis pipeline via a learned point cloud and a differentiable soft z-buffer method, where a point is projected to a region in the image plane with some radius using $\alpha$-compositing with other projected points (regions). 
Shih~\etal~\cite{shih20203dp} regarded the input depth image as a layered structure, and the learning-based inpainting model synthesizes color-and-depth content into the occluded region in a spatial context-aware manner. 
These works usually assume that the viewpoint changes are small, which makes it nearly impossible to directly employ them. 
On the other hand, synthesis and rendering with arbitrary viewpoint changes often achieved by multiple images input~\cite{sitzmann2019scene,thies2019deferred,sitzmann2019deepvoxels,meshry2019neural,mildenhall2020nerf}. 
Traditional methods usually adopt the image-based rendering technique \cite{shum2000review} to generate novel views. 
Riegler~\etal~\cite{Riegler2020FVS} employed differentiable reprojection of image features. Sitzmann~\etal~\cite{sitzmann2019deepvoxels} learned a 3D-structured scene representation from only 2D supervision that encodes the view-dependent appearance of a 3D scene. 
Sitzmann~\etal~\cite{sitzmann2019scene} further proposed a implicit 3D scene representation which could be also learned from 2D images via a differentiable ray-marching algorithm.  
Mildenhall~\etal~\cite{mildenhall2020nerf} propose to represent scenes as 5D neural radiance fields which could render photorealistic novel views of complex scenes.
Meshryet~\etal~\cite{meshry2019neural} uses additional depth and semantic information of the point cloud, together with an encoded latent vector to achieve realistic rendering with different styles.
Recent surveys on neural rendering can be found in \cite{tewari2020neuralrendering,kato2020differentiable}.
All these methods require a set of images or the built point cloud as input in order to learn detailed 3D scene representation with the deep network. 
Since our input is only a single satellite image, it is even more difficult for the network to learn meaningful representation.

\section{Method}

We introduce a novel framework for synthesizing street-view panoramic video from a single satellite image and provide an overview of our proposed pipeline in Fig.~\ref{fig:pipeline}.
As shown in the figure, we use a cascaded network architecture with three stages: a satellite stage, a transformation stage, and a 3D-to-video generation stage.
The satellite stage is similar to the current state-of-the-art method S2G~\cite{lu20geometry} and estimates both depth map and semantics from an input satellite image.
Different from the geo-transformation layer used in S2G~\cite{lu20geometry} which transforms the satellite domain to the street view, we directly extract visible points from the constructed occupancy grid according to the given input trajectory. % to ensure geometric consistency.
In the last 3D-to-video generation stage, two cascaded networks are utilized to generate a feature point cloud from semantics, followed by a projection to each video frame and a light-weight upsampling module.
The second and third stages are detailed in the following subsections.

%------------------------------------------------------------------------------
\subsection{Visible Points Extraction}\label{sec:extraction}
%------------------------------------------------------------------------------
%
We first build a semantic voxel occupancy grid using the depth and semantic images from the satellite stage.
Together with the sampling locations in the input trajectory, we create a point cloud with only visible points and build 3D-2D correspondences.
This corresponds to finding the index of the point in the 3D space for each pixel in the video. 
Each pixel has a uniquely corresponding 3D point, and each point in the 3D space may correspond to multiple pixels. The same mapping will also be utilized for projecting the colored point cloud onto the video frames in the final step of the 3D-to-video generation stage.

Algo.~\ref{algo:extraction} describes the detailed procedure for extracting visible points and building 3D-2D correspondences. 
The algorithm takes as input the voxelized occupancy grid $V$ and ordered sampling locations $L \in \mathbb{R}^{T \times 2}$. Here, $T$ denotes the number of sampling locations, which is equal to the number of video frames.
The final outputs consist of an ordered set $P_T$ saving the 3D coordinates $(x,y,z)$ of all visible points and a mapping tensor $M \in \mathbb{R}^{T \times H \times W}$ for all 2D frame pixels.
Each element $M_{tpq}$ keeps an index value $i$ if the frame pixel in position $(t,p,q)$ corresponds to the $i\text{-th}$ visible point in $P_T$.
The ordered set of visible points and mapping matrix are iteratively computed. 
We assign value of $0$ to all frame pixels that have no corresponding point in the point cloud $P_t$ in the current iteration.

At each time step $t$, we first obtain a dense depth map $d \in \mathbb{R}^{H\times W}$ for the frame at location $L_t$ by taking a $\mathrm{z-buffer}$ operation in the occupancy grid $V$.
This processing step is identical to the geo-transformation layer proposed in S2G~\cite{lu20geometry}.
Then, a preliminary mapping $m \in \{0,1,...,|P_{t-1}|\}^{H\times W}$, which indicates the correspondence between the current frame pixels and the visible points set $P_{t-1}$ so far, is calculated by the $\mathrm{project}$ function. 
$m_{pq} = i$ means that the $i\text{-th}$ point in $P_{t-1}$ is projected to the $(p,q)\text{-th}$ pixel \textbf{and} the depth value is in a range of $d_{pq}(1\pm\epsilon)$, otherwise $m_{pq} = 0$. 
In our experments, we set $\epsilon=0.5\%$. 
For pixels without corresponding points, \ie, $\{ (p,q) \mid m_{pq} = 0 \}$, we $\mathrm{unproject}$ them to the 3D space to obtain an additional ordered set $P_a$ containing the incremental visible points and an additional mapping $m_a$ saving the correspondences between these pixels and the incremental visible points.
It should be noted that the incremental indices satisfy $m_a \in \{0,|P_{t-1}|+1,...,|P_{t-1}|+|P_a|\}^{H\times W}$, where a pixel with a corresponding point in $P_{t-1}$ is assigned with 0 and pixels with correspondences in $P_a$ have the indices offset by $|P_{t-1}|$.
Hence, $m \odot m_a=\{0\}^{H\times W}$ always holds, where $\odot$ denotes the Hadamard product. 
Finally, we update the visible point set and mapping tensor by joining $P_{t-1}$ with $P_a$ and saving $m+m_a$ to $M_t$.

\vspace{-1em}

\begin{algorithm}[h!]
    \caption{Visible Points Extraction}
    \label{algo:extraction}
    \SetAlgoLined
    \SetKwInOut{Input}{Input}
    \SetKwInOut{Output}{Output}
    \SetKwInOut{Init}{Init}
    \Input{$V$ (occupancy grid), $L$ (locations)}
    \Output{$P_T$ (point cloud), $M$ (point-pixel mapping)}
    \Init{$P_0\gets\varnothing$, $M\gets\{0\}^{T \times H \times W}$}
    \For {$t\gets1$ \KwTo $T$} {
        $d\gets \mathrm{z-buffer} (V, L_t)$ \\
        $m\gets \mathrm{project} (P_{t-1}, L_t, d)$ \\
        $P_a, m_a\gets \mathrm{unproject} (L_t, d, m, |P_{t-1}|)$ \\
        $P_t \gets P_{t-1} \bigcup P_a$ \\
        $M_t\gets m + m_a$ \\
    }
\end{algorithm}

\vspace{-1.2em}
Since only the center frame has ground truth street-view RGB %,  in order to utilize this as much as possible, 
and to reflect the projection characteristics of the panorama image, the locations of the sampling points $L$ are inputted to the algorithm in an order of $c, c+1, c-1, c+2, c-2, ...$, where $c$ is the index of the center frame.

%------------------------------------------------------------------------------
\subsection{3D Generator}\label{sec:generator}
%------------------------------------------------------------------------------
%
In the 3D-to-video generation stage, we first infer features for the point cloud in the 3D space from the reprojected semantics.
The semantics of the points is gathered from the satellite semantics according to each point's coordinates in the horizontal plane. 
Distant points are simply labeled as \textit{sky}. 
The proposed 3D generator consists of a SparseConvNet~\cite{graham18sparse} and a RandLA-Net~\cite{hu20randlanet}, with a cascaded connection. 
Both networks are operating purely in the 3D domain and have an hourglass structure acting on coarse and fine generation successively.
Finally, the points are projected to frames, which are further turned into the output video via a light-weight upsampling module.

The \textbf{coarse generation stage} is based on voxels. 
At the beginning of this stage, the point cloud is first voxelized according to the targeted voxel size. 
Multiple points sharing the same voxel will be averaged as the feature of that voxel. 
In our experiments, the voxel size is set to 3.125cm (32 voxels per meter). 
The SparseConvNet~\cite{graham18sparse} only operates on the occupied area of the voxel grid avoiding unnecessary computations on free space and thus achieving time- and memory-efficient 3D convolutions. 
Finally, the output of the network is de-voxelized to a point cloud. 
Again, points sharing the same voxel will be assigned to the same feature. 
As depicted in Fig.~\ref{fig:pipeline}, the visualized point cloud with intermediate coarse features already shows some characteristics of the building facade like windows.

The \textbf{fine generation stage} is based on the point cloud.
The input of this stage is a concatenation of the intermediate coarse features and the original point semantics from the skip connection. 
RandLA-Net~\cite{hu20randlanet} is an efficient and light-weight state-of-the-art architecture designed for semantic segmentation of large-scale point clouds. 
We leverage this network to infer the fine features for each point.
We set the number of nearest neighbors to 8, and the decimation ratio in its local feature aggregation module to 4.

Each pixel in the video frame then gathers both coarse and fine features from its corresponding point in the point cloud according to the point-pixel mapping $M$ computed in the transformation stage. 
Finally, the \textbf{upsampling module} doubles the resolution and turns the frames with rich features into the output RGB video.
In order not to break the consistency from the 3D space, the module is designed only with very few parameters.

The reason for using a cascaded architecture of these two networks rather than only using RandLA-Net~\cite{hu20randlanet} is that its efficient setting makes the size of the network rather small, but the capacity may not be enough to support a scene generation. 
With the help of SparseConvNet~\cite{graham18sparse} which is good at learning high-level features, RandLA-Net is able to better infer fine features from local information. 
% The two networks compliment each other while performing separate duties. 
We also conduct experiments on a generator with only RandLA-Net~\cite{hu20randlanet} as detailed in Sec.~\ref{sec:ablation}.

%------------------------------------------------------------------------------
\subsection{Multi-class Encoder}\label{sec:encoder}
%------------------------------------------------------------------------------
%
S2G~\cite{lu20geometry} follows BicycleGAN~\cite{zhu2017toward} to use a single latent vector when generating the whole scene. Instead, we use a multi-class texture encoder that computes several latent vectors per class to enrich the diversity of generated scenes.

\begin{figure*}[t]
    \centering
    \footnotesize
    \newcommand{\sz}{0.075\linewidth} % figure size
    \newcommand{\zs}{0.15\linewidth}
    % \noindent\animategraphics[autoplay,loop,height=\sz]{5}{fig/forward/slt_sate_frames/16/}{00}{\NUMFRAMEA}
    % \noindent\animategraphics[autoplay,loop,height=\sz]{5}{fig/forward/slt_gt_frames/16/}{00}{\NUMFRAMEA}
    % \noindent\animategraphics[autoplay,loop,height=\sz]{5}{fig/forward/slt_s2g-f/16/}{00}{\NUMFRAMEA}
    % \noindent\animategraphics[autoplay,loop,height=\sz]{5}{fig/forward/slt_s2g-i/16/}{00}{\NUMFRAMEA}
    % \noindent\animategraphics[autoplay,loop,height=\sz]{5}{fig/forward/slt_vid2vid/16/}{00}{\NUMFRAMEA}
    % \noindent\animategraphics[autoplay,loop,height=\sz]{5}{fig/forward/slt_wc_vid2vid/16/}{00}{\NUMFRAMEA}
    % \noindent\animategraphics[autoplay,loop,height=\sz]{5}{fig/forward/slt_ours_jpg/16/}{00}{\NUMFRAMEA} \\ [1pt]
    
    \noindent\animategraphics[autoplay,loop,height=\sz]{5}{fig/forward/slt_sate_frames/33/}{00}{\NUMFRAMEA}
    \noindent\animategraphics[autoplay,loop,height=\sz]{5}{fig/forward/slt_gt_frames/33/}{00}{\NUMFRAMEA}
    \noindent\animategraphics[autoplay,loop,height=\sz]{5}{fig/forward/slt_s2g-f/33/}{00}{\NUMFRAMEA}
    \noindent\animategraphics[autoplay,loop,height=\sz]{5}{fig/forward/slt_s2g-i/33/}{00}{\NUMFRAMEA}
    \noindent\animategraphics[autoplay,loop,height=\sz]{5}{fig/forward/slt_vid2vid/33/}{00}{\NUMFRAMEA}
    \noindent\animategraphics[autoplay,loop,height=\sz]{5}{fig/forward/slt_wc_vid2vid/33/}{00}{\NUMFRAMEA}
    \noindent\animategraphics[autoplay,loop,height=\sz]{5}{fig/forward/slt_ours_jpg/33/}{00}{\NUMFRAMEA} \\ [1pt]
    
    \noindent\animategraphics[autoplay,loop,height=\sz]{5}{fig/forward/slt_sate_frames/82/}{00}{\NUMFRAMEA}
    \noindent\animategraphics[autoplay,loop,height=\sz]{5}{fig/forward/slt_gt_frames/82/}{00}{\NUMFRAMEA}
    \noindent\animategraphics[autoplay,loop,height=\sz]{5}{fig/forward/slt_s2g-f/82/}{00}{\NUMFRAMEA}
    \noindent\animategraphics[autoplay,loop,height=\sz]{5}{fig/forward/slt_s2g-i/82/}{00}{\NUMFRAMEA}
    \noindent\animategraphics[autoplay,loop,height=\sz]{5}{fig/forward/slt_vid2vid/82/}{00}{\NUMFRAMEA}
    \noindent\animategraphics[autoplay,loop,height=\sz]{5}{fig/forward/slt_wc_vid2vid/82/}{00}{\NUMFRAMEA}
    \noindent\animategraphics[autoplay,loop,height=\sz]{5}{fig/forward/slt_ours_jpg/82/}{00}{\NUMFRAMEA} \\ [1pt]
    
    % \noindent\animategraphics[autoplay,loop,height=\sz]{5}{fig/forward/slt_sate_frames/117/}{00}{\NUMFRAMEA}
    % \noindent\animategraphics[autoplay,loop,height=\sz]{5}{fig/forward/slt_gt_frames/117/}{00}{\NUMFRAMEA}
    % \noindent\animategraphics[autoplay,loop,height=\sz]{5}{fig/forward/slt_s2g-f/117/}{00}{\NUMFRAMEA}
    % \noindent\animategraphics[autoplay,loop,height=\sz]{5}{fig/forward/slt_s2g-i/117/}{00}{\NUMFRAMEA}
    % \noindent\animategraphics[autoplay,loop,height=\sz]{5}{fig/forward/slt_vid2vid/117/}{00}{\NUMFRAMEA}
    % \noindent\animategraphics[autoplay,loop,height=\sz]{5}{fig/forward/slt_wc_vid2vid/117/}{00}{\NUMFRAMEA}
    % \noindent\animategraphics[autoplay,loop,height=\sz]{5}{fig/forward/slt_ours_jpg/117/}{00}{\NUMFRAMEA} \\ [1pt]
    
    % \noindent\animategraphics[autoplay,loop,height=\sz]{5}{fig/forward/slt_sate_frames/162/}{00}{\NUMFRAMEA}
    % \noindent\animategraphics[autoplay,loop,height=\sz]{5}{fig/forward/slt_gt_frames/162/}{00}{\NUMFRAMEA}
    % \noindent\animategraphics[autoplay,loop,height=\sz]{5}{fig/forward/slt_s2g-f/162/}{00}{\NUMFRAMEA}
    % \noindent\animategraphics[autoplay,loop,height=\sz]{5}{fig/forward/slt_s2g-i/162/}{00}{\NUMFRAMEA}
    % \noindent\animategraphics[autoplay,loop,height=\sz]{5}{fig/forward/slt_vid2vid/162/}{00}{\NUMFRAMEA}
    % \noindent\animategraphics[autoplay,loop,height=\sz]{5}{fig/forward/slt_wc_vid2vid/162/}{00}{\NUMFRAMEA}
    % \noindent\animategraphics[autoplay,loop,height=\sz]{5}{fig/forward/slt_ours_jpg/162/}{00}{\NUMFRAMEA} \\ [1pt]
    
    \noindent\animategraphics[autoplay,loop,height=\sz]{5}{fig/forward/slt_sate_frames/163/}{00}{\NUMFRAMEA}
    \noindent\animategraphics[autoplay,loop,height=\sz]{5}{fig/forward/slt_gt_frames/163/}{00}{\NUMFRAMEA}
    \noindent\animategraphics[autoplay,loop,height=\sz]{5}{fig/forward/slt_s2g-f/163/}{00}{\NUMFRAMEA}
    \noindent\animategraphics[autoplay,loop,height=\sz]{5}{fig/forward/slt_s2g-i/163/}{00}{\NUMFRAMEA}
    \noindent\animategraphics[autoplay,loop,height=\sz]{5}{fig/forward/slt_vid2vid/163/}{00}{\NUMFRAMEA}
    \noindent\animategraphics[autoplay,loop,height=\sz]{5}{fig/forward/slt_wc_vid2vid/163/}{00}{\NUMFRAMEA}
    \noindent\animategraphics[autoplay,loop,height=\sz]{5}{fig/forward/slt_ours_jpg/163/}{00}{\NUMFRAMEA} \\ [1pt]
    
    \noindent\animategraphics[autoplay,loop,height=\sz]{5}{fig/forward/slt_sate_frames/199/}{00}{\NUMFRAMEA}
    \noindent\animategraphics[autoplay,loop,height=\sz]{5}{fig/forward/slt_gt_frames/199/}{00}{\NUMFRAMEA}
    \noindent\animategraphics[autoplay,loop,height=\sz]{5}{fig/forward/slt_s2g-f/199/}{00}{\NUMFRAMEA}
    \noindent\animategraphics[autoplay,loop,height=\sz]{5}{fig/forward/slt_s2g-i/199/}{00}{\NUMFRAMEA}
    \noindent\animategraphics[autoplay,loop,height=\sz]{5}{fig/forward/slt_vid2vid/199/}{00}{\NUMFRAMEA}
    \noindent\animategraphics[autoplay,loop,height=\sz]{5}{fig/forward/slt_wc_vid2vid/199/}{00}{\NUMFRAMEA}
    \noindent\animategraphics[autoplay,loop,height=\sz]{5}{fig/forward/slt_ours_jpg/199/}{00}{\NUMFRAMEA} \\ [1pt]

    \noindent\animategraphics[autoplay,loop,height=\sz]{5}{fig/forward/slt_sate_frames/284/}{00}{\NUMFRAMEA}
    \noindent\animategraphics[autoplay,loop,height=\sz]{5}{fig/forward/slt_gt_frames/284/}{00}{\NUMFRAMEA}
    \noindent\animategraphics[autoplay,loop,height=\sz]{5}{fig/forward/slt_s2g-f/284/}{00}{\NUMFRAMEA}
    \noindent\animategraphics[autoplay,loop,height=\sz]{5}{fig/forward/slt_s2g-i/284/}{00}{\NUMFRAMEA}
    \noindent\animategraphics[autoplay,loop,height=\sz]{5}{fig/forward/slt_vid2vid/284/}{00}{\NUMFRAMEA}
    \noindent\animategraphics[autoplay,loop,height=\sz]{5}{fig/forward/slt_wc_vid2vid/284/}{00}{\NUMFRAMEA}
    \noindent\animategraphics[autoplay,loop,height=\sz]{5}{fig/forward/slt_ours_jpg/284/}{00}{\NUMFRAMEA} \\ [-3pt]
    
    \setlength{\tabcolsep}{1pt}
    \begin{tabular}{C{\sz}C{\zs}C{\zs}C{\zs}C{\zs}C{\zs}C{\zs}}
    Sat. & Ground Truth & S2G-F~\cite{lu20geometry} & S2G-I~\cite{lu20geometry} & Vid2Vid~\cite{wang2018vid2vid} & WC-Vid2Vid~\cite{mallya2020world} & \OURS{} (\textbf{Ours}) \\ [-10pt]
    \end{tabular}
    \caption{\textbf{Qualitative comparison to baselines.} We show comparisons to state of the arts on a variety of test results. Our method generates significantly more realistic videos with better temporal consistency and contains fewer artifacts. \USEADOBE}
    \label{fig:eval}
    \vspace{-1.5em}
\end{figure*}

The encoder in the BicycleGAN~\cite{zhu2017toward} used in our pipeline takes as input the ground truth street-view RGB, as well as the semantics of the center frame during training.
The role of the semantics here is an indicator used for attentive pooling. 
After obtaining the feature map $F$ of the entire image, the encoder does not directly perform average pooling but instead pools the features of pixels with the same semantic class to finally obtain multiple latent vectors. 
For a specific class $c$, its corresponding semantic map $S^c$ is used for attentive pooling to finally obtain the latent vector $v_c$ of this class, \ie, $v_c = (\sum_{ij} S_{ij}^cF_{ij}) / (\sum_{ij} S_{ij}^c)$, where $i,j$ denote the spatial indices.
The encoder for the satellite image is similar to the encoder in the BicycleGAN. 
During training, the goal is to make the generated latent vectors as similar as possible to what is generated by the encoder in the BicycleGAN. 
Since some of the classes, \eg \textit{sky}, and \textit{sidewalk}, may not be able to infer from the satellite image, there is no loss on the latent vectors for these classes during training and they are directly given random vectors during inference.

\section{Experiments}

\begin{table*}[t]
    \centering
    \footnotesize
    \setlength{\tabcolsep}{13.6pt}
	\newcommand{\spacecell}{\hspace{9pt}-\hspace{9pt}}
	\newcommand{\largespacecell}{\hspace{11pt}-\hspace{11pt}}
	%normal font size
 	%\setlength{\tabcolsep}{7.15pt}
	%\newcommand{\spacecell}{\hspace{11pt}-\hspace{11pt}}
	%\newcommand{\largespacecell}{\hspace{13pt}-\hspace{13pt}}
	% \vspace{-10pt}
    % \scalebox{0.695}
    % \begin{adjustbox}{width=0.96\textwidth}
    % \renewcommand{\arraystretch}{1.2}
	\begin{tabular}{@{}l c c c c c c@{}}
	\toprule
	\multirow{1}{*}{\textbf{Method}} & PSNR$\uparrow$ & SSIM$\uparrow$ & Sharp Diff.$\uparrow$ & P$_{\text{Alex}}\downarrow$ & P$_{\text{Squeeze}}\downarrow$ & P$_{\text{VGG}}\downarrow$ \\

    \cmidrule(r){1-1} \cmidrule(lr){2-4} \cmidrule(l){5-7}
	Pix2Pix~\cite{isola2017image} & \largespacecell / 13.257 & \spacecell / 0.313 & \largespacecell / 24.673 & \spacecell / 0.606 & \spacecell / 0.478 & \spacecell / 0.629 \\
	Regmi~\etal~\cite{regmi2018cross} & \largespacecell / 13.305 & \spacecell / 0.320 & \largespacecell / 24.560 & \spacecell / 0.587 & \spacecell / 0.443 & \spacecell / 0.600 \\
	S2G-F~\cite{lu20geometry} & 14.110 / 14.146 & 0.347 / 0.346 & 25.851 / 25.861 & 0.530 / 0.528 & 0.422 / 0.422 & 0.626 / 0.626 \\
	S2G-I~\cite{lu20geometry} & 14.169 / 14.146 & 0.365 / 0.346 & \textbf{26.137} / 25.861 & 0.520 / 0.528 & 0.404 / 0.422 & 0.594 / 0.626  \\
	Vid2Vid~\cite{wang2018vid2vid} & 13.546 / 13.502 & 0.391 / 0.390 & 25.552 / 25.553 & 0.488 / 0.483 & 0.363 / 0.361 & 0.545 / 0.544 \\
	WC-Vid2Vid~\cite{mallya2020world} & 13.879 / 13.904 & 0.346 / 0.345 & 25.400 / 25.410 & 0.508 / 0.502 & 0.369 / 0.367 & 0.556 / 0.554 \\

    \cmidrule(r){1-1} \cmidrule(lr){2-4} \cmidrule(l){5-7}
    
	\OURS{} (\textbf{Ours}) & \textbf{15.171} / \textbf{15.220} & \textbf{0.409} / \textbf{0.410} & 26.068 / \textbf{26.060} & \textbf{0.482} / \textbf{0.478} & \textbf{0.342} / \textbf{0.342} & \textbf{0.535} / \textbf{0.533}  \\
	\bottomrule
    \end{tabular}
    % \end{adjustbox}
    \vspace{-7pt}
    \caption[]{\textbf{Quantitative evaluation of image quality}. For each entry we report two numbers indicating the evaluation on all frames and only on the center frame, respectively. Our method outperforms all baselines on most of the metrics.}
	\label{tab:eval}
	\vspace{-1em}
\end{table*}

%------------------------------------------------------------------------------
\subsection{Ground Truth}\label{sec:dataset}
%------------------------------------------------------------------------------
%
To the best of our knowledge, there is currently no available dataset that provides both satellite images and corresponding street-view panorama videos. 
As the first work that sheds light on the task of street-view video synthesis from a single satellite image, we first produce a dataset that satisfies the requirements for the task. Specifically, we extend the London panorama dataset used in S2G~\cite{lu20geometry} by generating the ground truth of street-view video snippets. 
The original dataset includes around 2K pairs of satellite images and corresponding street-view panoramas that are captured in the center position of the satellite images. 
The estimated depths (elevation) and semantics of the satellite image are also provided as ground truth.
% \sout{
% In brief, we generate the ground-truth street-view panorama videos by the interpolation in the 3D space via a point cloud, of which the geometry is calculated by the estimated depth of the street-view panorama.
% We elaborate on the details as follows.
% }
In brief, we interpolate the ground-truth street-view panorama videos in the 3D space via a point cloud, of which the geometry is calculated by the estimated depth of the available street-view panorama in the center position. 
We elaborate on the details as follows.
% We further provide a refined version by inferring dense depth from the street-view panorama.

\boldparagraph{Sampling trajectory.}
Each single street-view panorama image provided in the London panorama dataset~\cite{lu20geometry} is taken in the center of the satellite image and is associated with orientation.
To generate the street-view panorama video surrounding the location of this image, we set the sampling paths in both training and inference in a total range of 7 meters straight ahead and back from the viewing center. 
Taking the interval step of 0.5 meters, a total number of 15 frames including the center frame are sampled to form a video.
We denote the provided single street-view panorama image as the center frame for brevity.

\boldparagraph{Geometry.}
% \sout{
% Generating other frames by either the interpolation via a point cloud or simply warping requires precise enough geometry.
% However, due to the limited resolution, it is hard to infer accurate geometry from the satellite depth.
% That is the reason why we start directly from the street views.
% }
To generate panorama frames in novel positions, both the interpolation via a point cloud and simple warping require precise geometry of the scene. 
However, an accurate geometry is hard to be inferred from the satellite image considering its limited resolution and not accurate enough ground-truth elevation. 
Therefore, we infer the scene geometry from the available center frame instead of the satellite image.
We first generate a dense depth map for the center frame using MiDaS~\cite{ranftl20midas}, a state-of-the-art method for monocular depth estimation. % from single RGB images.
Although the pre-trained model used pinhole images, it still works well for panoramas.
We normalize the depth map by ensuring that the height of the viewing center (standing point) is 3 meters. Then we unproject the central frame depth to generate a raw 3D point cloud and obtain the depths for other frames by re-projecting the point cloud into each frame. 
For the location without a valid projection, we infer its missing depth value by exploiting the OpenCV inpainting function. 
Through unprojecting each frame's depth to the 3D space, a final point cloud can be constructed.

\boldparagraph{Interpolation via point cloud.}
Only points unprojected from the center frame possess the exact RGB information. For other points in the point cloud, we complement their colors through the nearest neighbor search. 
More specifically, for each uncolored point, we search for its 32 nearest-neighbor center-frame points that have valid information and determine its RGB by a distance-based weighted average on these neighbors respectively.
% A point cloud can be estimated from the satellite ground truth depth by using the algorithm described in Sec.~\ref{sec:extraction}. %
Finally, by re-projecting all colored points back to the frames, we can get a video of good quality.
The generated ground-truth video examples can be seen in Fig.~\ref{fig:eval}.

\boldparagraph{Semantics.}
Obtaining street-view semantic videos follows the procedure mentioned above.
We first adopt DeepLab v3+~\cite{chen18deeplabv3plus} with an Xception 71~\cite{chollet17xception} backbone and which is pre-trained on the Cityscapes~\cite{cordts16cityscapes} dataset to get the semantics of center frames.
Compared to SegNet~\cite{badri17segnet} utilized in S2G~\cite{lu20geometry}, DeepLab v3+ generates more accurate semantics.
The semantics of other frames are again complemented by the nearest neighbors search described above via a voting strategy instead of the weighted average used for RGB.

% Fig.~\ref{fig:gt_example} shows 3 examples of ground truth street-view panoramic frames generated based on ground truth satellite depth in S2G~\cite{lu20geometry} and the depth map predicted by MiDaS~\cite{ranftl20midas}. 
% The red bounding boxes clearly show that depth misalignments can cause shape distortions in the generated frame, which is alleviated by leveraging the refined point cloud.
% Since most RGB frames generated using satellite depth-based geometry are blurred or distorted, we adopt more consistent RGB frames with higher quality generated from the refined point cloud by using MiDaS~\cite{ranftl20midas} depth as the ground truth during the inference evaluation.
% In the training phase, we also use this as an incremental dataset. 
% As shown in the ablation study in Sec.~\ref{sec:ablation}, the incorporation of the incremental dataset improves the final performance.

%------------------------------------------------------------------------------
\subsection{Implementation Details}
%------------------------------------------------------------------------------
%
Our framework is implemented in PyTorch and run on a single Nvidia Tesla V100 GPU with 32GB memory. 
For the subnetworks, we use the official implementation of BicycleGAN~\cite{zhu2017toward} and SparseConvNet~\cite{graham18sparse}, as well as an unofficial RandLA-Net~\cite{hu20randlanet} PyTorch implementation\footnote{\url{https://github.com/aRI0U/RandLA-Net-pytorch}}.
For the dataset, we keep an output resolution of 512$\times$256 such that the point cloud size of each scene is around 200K.
During training, we use the geometry from the satellite depth to be consistent with the inference stage.
For the network architecture, the default training settings of BicycleGAN~\cite{zhu2017toward} are employed, using 16 for the size of latent vectors and 64 for the size of intermediate features.
% The discriminator uses all predicted frames as the fake set and RGBs from MiDaS~\cite{ranftl20midas} version as the true set.
The multi-noise encoder only takes as input the center frame. 
We further distinguish between left and right buildings in the semantic labels to achieve better diversity.
For the 3D generator, we use the default provided U-Net~\cite{ronneberger15unet} implementations under SparseConvNet~\cite{graham18sparse} and RandLA-Net~\cite{hu20randlanet} frameworks which are originally used for point cloud semantic segmentation.
% During training, the 3D generator loss is computed directly on the point cloud instead of being projected to each frame. If the point cloud geometry comes from the satellite stage.

Because the accuracy of the satellite depth is barely high enough, it is hard to ensure that the point cloud estimated from the satellite depth is well aligned with the generated ground-truth panorama video.
For instance, the distant points (geometrically \textit{sky} points) might be assigned with colors or semantics of the building if the estimated height of that building does not agree with the panorama. 
Especially, for the objects like street lamps and cars which are nearly invisible in the satellite image, misalignment is inevitable. 
% Such misalignments between color/semantics and geometry will result in distorted panorama frames. 
% Therefore, to mitigate such errors, we introduce an incremental training set by refining the geometry of the point cloud.
We simply use the semantic information to determine misalignments for which pixels will be given zero weights when computing the loss.
%\eg, distant points with a non-\textit{sky} label or nearby points having a \textit{sky} label

\begin{table*}[t]
    \centering
    \footnotesize
    \setlength{\tabcolsep}{12.5pt}
    % \scalebox{0.695}
    % \begin{adjustbox}{width=0.96\textwidth}
    % \renewcommand{\arraystretch}{1.1}
	\begin{tabular}{@{}l c c c c c c c c @{}}
	\toprule
	\multirow{1}{*}{\textbf{Method}} & MSE$_{\text{RGB}}\downarrow$ & PSNR$\uparrow$ & SSIM$\uparrow$ & Sharp Diff.$\uparrow$ & P$_{\text{Alex}}\downarrow$ & P$_{\text{Squeeze}}\downarrow$ & P$_{\text{VGG}}\downarrow$ & User Study \\

    \cmidrule(r){1-1} \cmidrule(lr){2-2} \cmidrule(lr){3-5} \cmidrule(lr){6-8} \cmidrule(l){9-9}
	Vid2Vid~\cite{wang2018vid2vid} & 
    21.605 & 21.764 & 0.774 & 30.950 & 0.116 & 0.077 & 0.211 & \textcolor{white}{0}9.3\%
	\\
	WC-Vid2Vid~\cite{mallya2020world} &
	10.604 & 27.783 & 0.871 & 35.296 & 0.108 & 0.074 & 0.176 & 32.6\%
	\\
    \cmidrule(r){1-1} \cmidrule(lr){2-2} \cmidrule(lr){3-5} \cmidrule(lr){6-8} \cmidrule(l){9-9}
	\OURS{} (\textbf{Ours}) & \textcolor{white}{0}\textbf{1.668} & \textbf{43.982} & \textbf{0.997} & \textbf{50.748} & \textbf{0.006} & \textbf{0.007} & \textbf{0.021} & \textbf{58.1\%}
	\\
	\bottomrule
    \end{tabular}
    % \end{adjustbox}
    \vspace{-7pt}
    \caption[]{\textbf{Quantitative evaluation of temporal self-consistency.} The evaluation is based on a u-turn-shaped trajectory.}
	\label{tab:uturn}
	\vspace{-1.5em}
\end{table*}

%------------------------------------------------------------------------------
\subsection{Baseline Comparison}
%------------------------------------------------------------------------------

Since we are the first to propose a method for generating street-view panoramic videos from single satellite images, we design two baseline methods by adapting the state-of-the-art street-view panoramic image synthesis method S2G~\cite{lu20geometry} for video generation:
\textbf{(1) S2G-F}: each frame is generated individually but shares the same latent vector encoded from the input satellite image; 
\textbf{(2) S2G-I}: only the center frame is generated and other frames are interpolated by using the point cloud coloring procedure described in Sec.~\ref{sec:dataset}. 
Vid2Vid~\cite{wang2018vid2vid} and WC-Vid2Vid~\cite{mallya2020world} are also included in the comparison.
As they are originally designed for video-to-video translation they cannot be directly employed.
We first generate per-frame semantics and pixel correspondences (only for WC-Vid2Vid~\cite{mallya2020world}) from the output of our satellite stage, which are required as input by these two methods.
%
%We compare our method with these baseline methods on the test set of London panorama~\cite{lu20geometry}.
The comparison is conducted on the test set of London panorama~\cite{lu20geometry}.

For quantitative evaluation, we follow \cite{lu20geometry} and use PSNR, SSIM, and sharpness difference (Sharp Diff.) as low-level metrics to measure the per-pixel differences between the predicted frames and the ground-truth video. 
The high-level perceptual similarity is also taken into account. 
P$_\text{Alex}$, P$_\text{Squeeze}$, P$_\text{VGG}$ denote the evaluation results based on the backbone of AlexNet~\cite{krizhevsky2012imagenet}, SqueezeNet~\cite{squeezenet} and VGG~\cite{vgg}, respectively.

In addition to the above two baselines, we compare to two image-to-image translation works, Pix2Pix~\cite{isola2017image} and Regmi~\cite{regmi2018cross}, on the center frame generation. 
The quantitative results are shown in Tab.~\ref{tab:eval}.
For the video generation comparison, our improved performance may result from the better temporal consistency of our generated video, since all methods use the same geometry inferred from the input satellite image.
Regarding the center frame comparison, we outperform all state-of-the-art methods on all metrics, which indicates superiority of our method in generating geometrically consistent single street-view panorama.

More qualitative results are presented in Fig.~\ref{fig:eval}. 
We can see that the frames generated by our method are both temporally and geometrically consistent. 
Since each frame from the baseline method S2G-F~\cite{lu20geometry} is synthesized independently, the textures in different frames are nearly stationary and there is no consistent transition between them when the observation location changes. 
Vid2Vid~\cite{wang2018vid2vid} has better per-frame appearance but still suffers from the problem of stationary patterns. 
This may be due to an inaccurate optical flow estimation within their network.
For S2G-I~\cite{lu20geometry}, we can see that the interpolation can ensure consistency of the texture between frames since every frame's texture comes from the center frame and is based on the geometry. 
Nevertheless, it is easy to find that the texture in the frames which are far away from the center frame is likely to be blurred, especially on the building facades which are invisible in the center frame.
WC-Vid2Vid~\cite{mallya2020world} generally have good consistency since pixel correspondences are provided as input. 
However, their appearances, especially the building facades, look similar across different examples.

\begin{figure}[t]
    \centering
    \footnotesize
    \newcommand{\sz}{0.139\columnwidth} % figure size
    \newcommand{\zs}{0.28\columnwidth}
    \vspace{0.3em}
    \noindent\animategraphics[autoplay,loop,height=\sz]{5}{fig/uturn/uturn_sate_frames/162/}{00}{\NUMFRAMEB}
    \noindent\animategraphics[autoplay,loop,height=\sz]{5}{fig/uturn/uturn_vid2vid/162/}{00}{\NUMFRAMEB}
    \noindent\animategraphics[autoplay,loop,height=\sz]{5}{fig/uturn/uturn_wc_vid2vid/162/}{00}{\NUMFRAMEB}
    \noindent\animategraphics[autoplay,loop,height=\sz]{5}{fig/uturn/uturn_ours_jpg/162/}{00}{\NUMFRAMEB} \\ [1pt]
    \noindent\animategraphics[autoplay,loop,height=\sz]{5}{fig/uturn/uturn_sate_frames/228/}{00}{\NUMFRAMEB}
    \noindent\animategraphics[autoplay,loop,height=\sz]{5}{fig/uturn/uturn_vid2vid/228/}{00}{\NUMFRAMEB}
    \noindent\animategraphics[autoplay,loop,height=\sz]{5}{fig/uturn/uturn_wc_vid2vid/228/}{00}{\NUMFRAMEB}
    \noindent\animategraphics[autoplay,loop,height=\sz]{5}{fig/uturn/uturn_ours_jpg/228/}{00}{\NUMFRAMEB} \\ [1pt]
    \noindent\animategraphics[autoplay,loop,height=\sz]{5}{fig/uturn/uturn_sate_frames/267/}{00}{\NUMFRAMEB}
    \noindent\animategraphics[autoplay,loop,height=\sz]{5}{fig/uturn/uturn_vid2vid/267/}{00}{\NUMFRAMEB}
    \noindent\animategraphics[autoplay,loop,height=\sz]{5}{fig/uturn/uturn_wc_vid2vid/267/}{00}{\NUMFRAMEB}
    \noindent\animategraphics[autoplay,loop,height=\sz]{5}{fig/uturn/uturn_ours_jpg/267/}{00}{\NUMFRAMEB} \\ [-3pt]
    \setlength{\tabcolsep}{1pt}
    \begin{tabular}{C{\sz}C{\zs}C{\zs}C{\zs}C{\zs}C{\zs}}
      Sat. & Vid2Vid & WC-Vid2Vid & Ours \\[-10pt]
    \end{tabular}
    \caption{\textbf{Qualitative evalutation of the self-temporal consistency.} Results are generated based on a u-turn-shaped trajectory. \USEADOBE}
    \label{fig:uturn}
    \vspace{-1.5em}
\end{figure}

%------------------------------------------------------------------------------
\subsection{Self-consistency Evaluation}\label{sec:consistency}
%------------------------------------------------------------------------------

To evaluate the temporal consistency of synthesized video frames between different methods, we designed an experiment based on a special u-turn-shaped trajectory, with a total of 60 frames.
We then compute the pixel-wise difference between the two frames of the same position in the two directions (directions are adjusted when computing the metric).
Such an evaluation is devised to assess the frame's temporal self-consistency in one consecutive synthesis.
Beside the metric used in Tab.~\ref{tab:eval}, we also compared the MSE value of RGB.

In addition, we conducted a user study, where we provided randomly selected 15 samples (including 10 forward motions and 5 u-turns) with results of Vid2Vid~\cite{wang2018vid2vid}, WC-Vid2Vid~\cite{mallya2020world}, and ours.
We asked 28 people to select only one result of the best naturalness and consistency for each sample, in a total of 420 votes.
All evaluations of the temporal self-consistency as well as the voting ratios of user study are detailed in Tab.~\ref{tab:uturn}.
We also present the synthesis results of the u-turn trajectory in Fig.~\ref{fig:uturn}. 
Both the quantitative and qualitative results indicate our method has significantly better self-consistency across frames than the two strong baseline methods.

%------------------------------------------------------------------------------
\subsection{Ablation Study}\label{sec:ablation}
%------------------------------------------------------------------------------

\begin{table*}[tb]
    \centering
    \footnotesize
    \setlength{\tabcolsep}{13.8pt}
    %normalsize
    %\setlength{\tabcolsep}{6.75pt}
    % \vspace{-15pt}
	\hspace{-2pt}
    % \scalebox{0.695}
    % \begin{adjustbox}{width=0.96\textwidth}
    % \renewcommand{\arraystretch}{1.1}
	\begin{tabular}{@{}l c c c c c c@{}}
	\toprule
	\textbf{Method} & PSNR$\uparrow$ & SSIM$\uparrow$ & Sharp Diff.$\uparrow$ & P$_{\text{Alex}}\downarrow$ & P$_{\text{Squeeze}}\downarrow$ & P$_{\text{VGG}}\downarrow$ \\

    \cmidrule(r){1-1} \cmidrule(lr){2-4} \cmidrule(l){5-7}
	R &
    13.686 / 13.739 & \textbf{0.417} / \textbf{0.417} & 25.726 / 25.736 & 0.584 / 0.580 & 0.443 / 0.443 & 0.621 / 0.619
	\\
	R+S &
	14.551 / 14.590 & 0.402 / 0.403 & 25.493 / 25.479 & 0.561 / 0.564 & 0.404 / 0.402 & 0.572 / 0.568
	\\
	R+S+M &
	14.655 / 14.714 & 0.385 / 0.391 & 25.811 / 25.823 & 0.551 / 0.546 & 0.403 / 0.399 & 0.576 / 0.572
	\\
	\cmidrule(r){1-1} \cmidrule(lr){2-4} \cmidrule(l){5-7}
	R+S+M+U (\textbf{Ours}) & 
	\textbf{15.171} / \textbf{15.220} & 0.409 / 0.410 & 26.068 / 26.060 & \textbf{0.482} / \textbf{0.478} & \textbf{0.342} / \textbf{0.342} & \textbf{0.535} / \textbf{0.533}
	\\
	R+S+M+U+W &
	14.546 / 14.576 & 0.394 / 0.394 & \textbf{26.341} / \textbf{26.349} & 0.503 / 0.500 & 0.345 / 0.346 & 0.541 / 0.539
	\\
	\bottomrule
    \end{tabular}
    % \end{adjustbox}
    \vspace{-7pt}
    \caption[]{\textbf{Quantitative ablation study.} For each method we report two numbers indicating evaluation on all frames and only on the center frame, respectively. In short, the ablations are: \textbf{R}: basic framework with RandLA-Net~\cite{hu20randlanet}; \textbf{+S}: adding SparseConvNet~\cite{graham18sparse}; \textbf{+M}: multi-noise encoder; \textbf{+U}: upsampling module; \textbf{+W}: warped satellite information.}
	\label{tab:ablation}
	\vspace{-0.5em}
\end{table*}

\begin{figure*}
	\centering
	\footnotesize
    \newcommand{\sz}{0.075\linewidth} % figure size
    \newcommand{\zs}{0.15\linewidth}
	
    \noindent\animategraphics[autoplay,loop,height=\sz]{5}{fig/forward/slt_sate_frames/228/}{00}{\NUMFRAMEA}
    \noindent\animategraphics[autoplay,loop,height=\sz]{5}{fig/forward/slt_gt_frames/228/}{00}{\NUMFRAMEA}
    \noindent\animategraphics[autoplay,loop,height=\sz]{5}{fig/ablation/R/228/}{00}{\NUMFRAMEA}
    \noindent\animategraphics[autoplay,loop,height=\sz]{5}{fig/ablation/RS/228/}{00}{\NUMFRAMEA}
    \noindent\animategraphics[autoplay,loop,height=\sz]{5}{fig/ablation/RSM/228/}{00}{\NUMFRAMEA}
    \noindent\animategraphics[autoplay,loop,height=\sz]{5}{fig/forward/slt_ours_jpg/228/}{00}{\NUMFRAMEA}
    \noindent\animategraphics[autoplay,loop,height=\sz]{5}{fig/ablation/RSMUW/228/}{00}{\NUMFRAMEA} \\[1pt]
    
    \noindent\animategraphics[autoplay,loop,height=\sz]{5}{fig/forward/slt_sate_frames/267/}{00}{\NUMFRAMEA}
    \noindent\animategraphics[autoplay,loop,height=\sz]{5}{fig/forward/slt_gt_frames/267/}{00}{\NUMFRAMEA}
    \noindent\animategraphics[autoplay,loop,height=\sz]{5}{fig/ablation/R/267/}{00}{\NUMFRAMEA}
    \noindent\animategraphics[autoplay,loop,height=\sz]{5}{fig/ablation/RS/267/}{00}{\NUMFRAMEA}
    \noindent\animategraphics[autoplay,loop,height=\sz]{5}{fig/ablation/RSM/267/}{00}{\NUMFRAMEA}
    \noindent\animategraphics[autoplay,loop,height=\sz]{5}{fig/forward/slt_ours_jpg/267/}{00}{\NUMFRAMEA}
    \noindent\animategraphics[autoplay,loop,height=\sz]{5}{fig/ablation/RSMUW/267/}{00}{\NUMFRAMEA} \\[1pt]
    
    \noindent\animategraphics[autoplay,loop,height=\sz]{5}{fig/forward/slt_sate_frames/284/}{00}{\NUMFRAMEA}
    \noindent\animategraphics[autoplay,loop,height=\sz]{5}{fig/forward/slt_gt_frames/284/}{00}{\NUMFRAMEA}
    \noindent\animategraphics[autoplay,loop,height=\sz]{5}{fig/ablation/R/284/}{00}{\NUMFRAMEA}
    \noindent\animategraphics[autoplay,loop,height=\sz]{5}{fig/ablation/RS/284/}{00}{\NUMFRAMEA}
    \noindent\animategraphics[autoplay,loop,height=\sz]{5}{fig/ablation/RSM/284/}{00}{\NUMFRAMEA}
    \noindent\animategraphics[autoplay,loop,height=\sz]{5}{fig/forward/slt_ours_jpg/284/}{00}{\NUMFRAMEA}
    \noindent\animategraphics[autoplay,loop,height=\sz]{5}{fig/ablation/RSMUW/284/}{00}{\NUMFRAMEA} \\ [-7pt]
    
	\setlength{\tabcolsep}{1pt}
    \begin{tabular}{C{\sz}C{\zs}C{\zs}C{\zs}C{\zs}C{\zs}C{\zs}}
    Sat. & Ground Truth & R & R+S & R+S+M & R+S+M+U (\textbf{Ours}) & R+S+M+U+W \\[-15pt]
    \end{tabular}
	\vspace{-7pt}
	
	\caption{\textbf{Qualitative ablation study.} We present exemplary qualitative results for various ablations of our method. 
	The synthesized images visibly contain more details and achieve higher levels realism with our full method. \USEADOBE
	}
	\vspace{-1.0em}
	\label{fig:ablation}
\end{figure*}

\begin{figure}[tb]
    \centering
    \footnotesize
    \newcommand{\sz}{0.195\columnwidth} % figure size
    \newcommand{\zs}{0.39\columnwidth} % figure size

    \includegraphics[height=\sz]{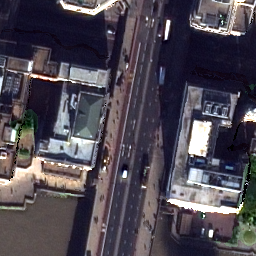}
    \includegraphics[height=\sz]{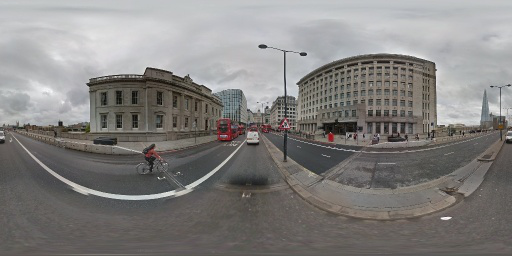}
    \includegraphics[height=\sz]{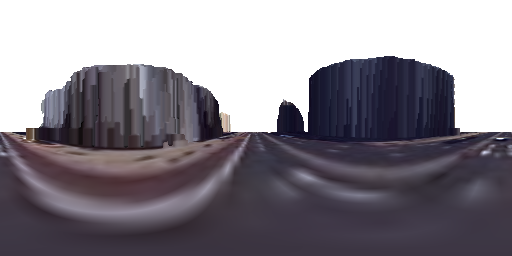}\\[1pt]
    \includegraphics[height=\sz]{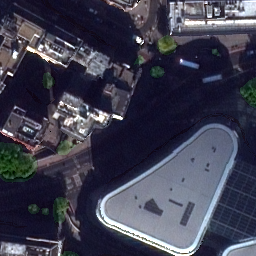}
    \includegraphics[height=\sz]{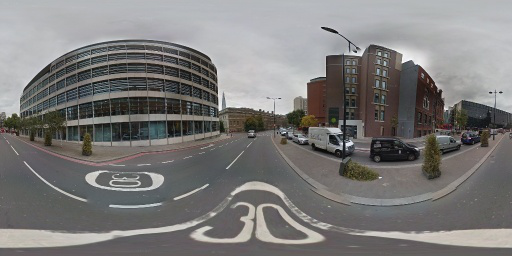}
    \includegraphics[height=\sz]{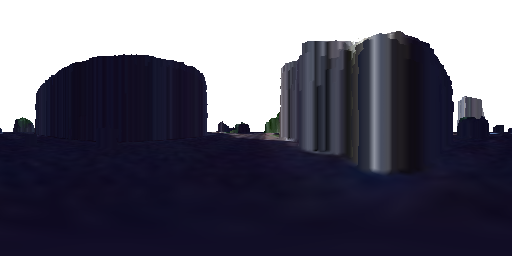}\\[-3pt]
    
    \setlength{\tabcolsep}{1pt}
    \begin{tabular}{C{\sz}C{\zs}C{\zs}}
      Sat. & Ground Truth & Warped Sat.\\[-10pt]
    \end{tabular}

    \caption{\textbf{Warped color satellite information.} The examples illustrate the low-quality of warped satellite images which often do not provide useful color information.}
    \label{fig:warp}
    \vspace{-1.5em}
\end{figure}

To better evaluate the effectiveness of the individual components of our method, we also conduct an ablation study by incrementally adding components into our basic framework. 
More specifically, we focus on the following three components: \textbf{(1)} the SparseConvNet~\cite{graham18sparse} used in the 3D generator; \textbf{(2)} the setting of multiple latent vectors; \textbf{(3)} the final upsampling module. 
We set the basic framework as the pipeline with only RandLA-Net~\cite{hu20randlanet} in the 3D generation stage, while our method possesses all components.

Tab.~\ref{tab:ablation} shows quantitative evaluation results of the ablation study.
The abbreviations of the method names in the table are defined as follows. \textbf{R}: the basic framework that uses RandLA-Net~\cite{hu20randlanet} and a global latent vector in the 3D generation stage; \textbf{R+S}: the coarse and fine generation framework by further incorporating SparseConvNet~\cite{graham18sparse}; \textbf{R+S+M}: further using a multi-class encoder to the R+S setting. \textbf{R+S+M+U}: further adding the upsampling module which forms our final method with all components.

The effectiveness of each added component is shown by clear performance improvements of the PSNR, P$_\text{Alex}$, and P$_\text{Squeeze}$ metrics.
Fig.~\ref{fig:ablation} further shows a qualitative comparison of the results generated by the aforementioned methods. 
As illustrated, frames generated by the full framework show higher consistency and smoothness over time compared with the other ablation variants.
Especially, the addition of SparseConvNet~\cite{graham18sparse} \textbf{(R+S)} significantly improves the generation quality compared to the basic setting \textbf{(R)} that only uses RandLA-Net~\cite{hu20randlanet}, which can only give the overall color and cannot restore the texture details, \eg, the building facade.
We address the main reason as the explicit allocation of the coarse generation and fine generation to two cascaded different networks respectively. 
This alleviates the struggle of RandLA-Net~\cite{hu20randlanet} in generating both coarse and fine textures.
With the introduction of multi-class encoder that generates multiple latent vectors \textbf{(R+S+M)}, the performance is further improved since it disentangles the latent vectors for different classes and enables more generation possibilities.
The upsampling module (\textbf{R+S+M+U}) further double the output resolution and makes the frames much clearer and realistic.

We also tried direct warping of satellite images as a part of the input of the upsampling module to better utilize the input information.
However, the evaluation of this addition (\textbf{R+S+M+U+W}) yields lower PSNR metrics, as well as perceptual similarity and SSIM scores, which indicates that the warped satellite images do not directly provide useful information.
This is might due to the limited resolution but also due to artifacts like cast shadows which make it difficult to readily extract useful color information.
Fig.~\ref{fig:warp} shows two examples of warped satellite images.
In a few cases like the first example, the road and the lane line can be warped into the street view, although very blurry.
In many cases like the second one, the road is covered by shadows of neighboring buildings resulting in a very dark the warping result.
This also illustrates the difficulty of cross-view video generation from a single satellite image.

\section{Conclusion}
We proposed a novel approach for cross-view video synthesis.
In particular, we presented a multi-stage pipeline that takes as input a single satellite image with a given trajectory, and generates a street-view panoramic video with both geometrical and temporal consistency.
The generator in our pipeline uses a point cloud, which is built from the input satellite image, as the basic structure and adopts a successively coarse and fine generation estimated by two cascaded hourglass structures.
The main generation in a 3D manner with a follow-up light-weight upsampling module significantly improves both the geometrical and temporal consistency across frames. % which is especially important for video synthesis.
Our experiments demonstrate that our method outperforms existing state-of-the-art cross-view generation or video translation approaches and is able to synthesize more realistic street-view panoramic videos and in larger variability.
To the best of our knowledge, we presented the first work that synthesizes videos under cross-view settings.

\clearpage

{
\small
\bibliographystyle{ieee_fullname}
\bibliography{sat2vid}

\begin{thebibliography}{10}\itemsep=-1pt

\bibitem{badri17segnet}
Vijay Badrinarayanan, Alex Kendall, and Roberto Cipolla.
\newblock Segnet: A deep convolutional encoder-decoder architecture for image
  segmentation.
\newblock {\em IEEE Transactions on Pattern Analysis and Machine Intelligence
  (TPAMI)}, 2017.

\bibitem{chan2019everybody}
Caroline Chan, Shiry Ginosar, Tinghui Zhou, and Alexei~A Efros.
\newblock Everybody dance now.
\newblock In {\em Proceedings of the IEEE International Conference on Computer
  Vision}, pages 5933--5942, 2019.

\bibitem{chen18deeplabv3plus}
Liang-Chieh Chen, Yukun Zhu, George Papandreou, Florian Schroff, and Hartwig
  Adam.
\newblock Encoder-decoder with atrous separable convolution for semantic image
  segmentation.
\newblock In {\em ECCV}, 2018.

\bibitem{chen2019mocycle}
Yang Chen, Yingwei Pan, Ting Yao, Xinmei Tian, and Tao Mei.
\newblock Mocycle-gan: Unpaired video-to-video translation.
\newblock In {\em Proceedings of the 27th ACM International Conference on
  Multimedia}, pages 647--655, 2019.

\bibitem{chollet17xception}
Francois Chollet.
\newblock Xception: Deep learning with depthwise separable convolutions.
\newblock In {\em Proceedings of the IEEE Conference on Computer Vision and
  Pattern Recognition (CVPR)}, July 2017.

\bibitem{cordts16cityscapes}
Marius Cordts, Mohamed Omran, Sebastian Ramos, Timo Rehfeld, Markus Enzweiler,
  Rodrigo Benenson, Uwe Franke, Stefan Roth, and Bernt Schiele.
\newblock The cityscapes dataset for semantic urban scene understanding.
\newblock In {\em Proceedings of the IEEE Conference on Computer Vision and
  Pattern Recognition (CVPR)}, June 2016.

\bibitem{finn2016unsupervised}
Chelsea Finn, Ian Goodfellow, and Sergey Levine.
\newblock Unsupervised learning for physical interaction through video
  prediction.
\newblock {\em arXiv preprint arXiv:1605.07157}, 2016.

\bibitem{goodfellow2014generative}
Ian Goodfellow, Jean Pouget-Abadie, Mehdi Mirza, Bing Xu, David Warde-Farley,
  Sherjil Ozair, Aaron Courville, and Yoshua Bengio.
\newblock Generative adversarial nets.
\newblock In {\em Advances in Neural Information Processing Systems}, pages
  2672--2680, 2014.

\bibitem{graham18sparse}
Benjamin Graham, Martin Engelcke, and Laurens van~der Maaten.
\newblock 3d semantic segmentation with submanifold sparse convolutional
  networks.
\newblock In {\em Proceedings of the IEEE Conference on Computer Vision and
  Pattern Recognition (CVPR)}, June 2018.

\bibitem{hao2018controllable}
Zekun Hao, Xun Huang, and Serge Belongie.
\newblock Controllable video generation with sparse trajectories.
\newblock In {\em Proceedings of the IEEE Conference on Computer Vision and
  Pattern Recognition}, pages 7854--7863, 2018.

\bibitem{hu20randlanet}
Qingyong Hu, Bo Yang, Linhai Xie, Stefano Rosa, Yulan Guo, Zhihua Wang, Niki
  Trigoni, and Andrew Markham.
\newblock Randla-net: Efficient semantic segmentation of large-scale point
  clouds.
\newblock In {\em Proceedings of the IEEE/CVF Conference on Computer Vision and
  Pattern Recognition (CVPR)}, June 2020.

\bibitem{squeezenet}
Forrest~N Iandola, Song Han, Matthew~W Moskewicz, Khalid Ashraf, William~J
  Dally, and Kurt Keutzer.
\newblock Squeezenet: Alexnet-level accuracy with 50x fewer parameters and< 0.5
  mb model size.
\newblock {\em arXiv preprint arXiv:1602.07360}, 2016.

\bibitem{isola2017image}
Phillip Isola, Jun-Yan Zhu, Tinghui Zhou, and Alexei~A Efros.
\newblock Image-to-image translation with conditional adversarial networks.
\newblock In {\em Proceedings of the IEEE Conference on Computer Vision and
  Pattern Recognition}, pages 1125--1134, 2017.

\bibitem{kato2020differentiable}
Hiroharu Kato, Deniz Beker, Mihai Morariu, Takahiro Ando, Toru Matsuoka, Wadim
  Kehl, and Adrien Gaidon.
\newblock Differentiable rendering: A survey, 2020.

\bibitem{krizhevsky2012imagenet}
Alex Krizhevsky, Ilya Sutskever, and Geoffrey~E Hinton.
\newblock Imagenet classification with deep convolutional neural networks.
\newblock In {\em Advances in neural information processing systems}, pages
  1097--1105, 2012.

\bibitem{li2018flow}
Yijun Li, Chen Fang, Jimei Yang, Zhaowen Wang, Xin Lu, and Ming-Hsuan Yang.
\newblock Flow-grounded spatial-temporal video prediction from still images.
\newblock In {\em Proceedings of the European Conference on Computer Vision
  (ECCV)}, pages 600--615, 2018.

\bibitem{liang2017dual}
Xiaodan Liang, Lisa Lee, Wei Dai, and Eric~P Xing.
\newblock Dual motion gan for future-flow embedded video prediction.
\newblock In {\em Proceedings of the IEEE International Conference on Computer
  Vision}, pages 1744--1752, 2017.

\bibitem{logacheva2020deeplandscape}
Elizaveta Logacheva, Roman Suvorov, Oleg Khomenko, Anton Mashikhin, and Victor
  Lempitsky.
\newblock Deeplandscape: Adversarial modeling of landscape videos.
\newblock In {\em Proceedings of the European Conference on Computer Vision
  (ECCV)}, August 2020.

\bibitem{lotter2016deep}
William Lotter, Gabriel Kreiman, and David Cox.
\newblock Deep predictive coding networks for video prediction and unsupervised
  learning.
\newblock {\em arXiv preprint arXiv:1605.08104}, 2016.

\bibitem{lu20geometry}
Xiaohu Lu, Zuoyue Li, Zhaopeng Cui, Martin~R. Oswald, Marc Pollefeys, and
  Rongjun Qin.
\newblock Geometry-aware satellite-to-ground image synthesis for urban areas.
\newblock In {\em Proceedings of the IEEE/CVF Conference on Computer Vision and
  Pattern Recognition (CVPR)}, June 2020.

\bibitem{mallya2020world}
Arun Mallya, Ting-Chun Wang, Karan Sapra, and Ming-Yu Liu.
\newblock World-consistent video-to-video synthesis.
\newblock In {\em Eur. Conf. Comput. Vis.}, 2020.

\bibitem{mathieu2015deep}
Michael Mathieu, Camille Couprie, and Yann LeCun.
\newblock Deep multi-scale video prediction beyond mean square error.
\newblock {\em arXiv preprint arXiv:1511.05440}, 2015.

\bibitem{meshry2019neural}
Moustafa Meshry, Dan~B Goldman, Sameh Khamis, Hugues Hoppe, Rohit Pandey, Noah
  Snavely, and Ricardo Martin-Brualla.
\newblock Neural rerendering in the wild.
\newblock In {\em Proceedings of the IEEE Conference on Computer Vision and
  Pattern Recognition}, pages 6878--6887, 2019.

\bibitem{mildenhall2020nerf}
Ben Mildenhall, Pratul~P Srinivasan, Matthew Tancik, Jonathan~T Barron, Ravi
  Ramamoorthi, and Ren Ng.
\newblock Nerf: Representing scenes as neural radiance fields for view
  synthesis.
\newblock In {\em Eur. Conf. Comput. Vis.}, 2020.

\bibitem{pan2019video}
Junting Pan, Chengyu Wang, Xu Jia, Jing Shao, Lu Sheng, Junjie Yan, and
  Xiaogang Wang.
\newblock Video generation from single semantic label map.
\newblock In {\em Proceedings of the IEEE Conference on Computer Vision and
  Pattern Recognition}, pages 3733--3742, 2019.

\bibitem{ranftl20midas}
Ren\'{e} Ranftl, Katrin Lasinger, David Hafner, Konrad Schindler, and Vladlen
  Koltun.
\newblock Towards robust monocular depth estimation: Mixing datasets for
  zero-shot cross-dataset transfer.
\newblock {\em IEEE Transactions on Pattern Analysis and Machine Intelligence
  (TPAMI)}, 2020.

\bibitem{regmi2018cross}
Krishna Regmi and Ali Borji.
\newblock Cross-view image synthesis using conditional gans.
\newblock In {\em Proceedings of the IEEE Conference on Computer Vision and
  Pattern Recognition}, pages 3501--3510, 2018.

\bibitem{regmi2019bridging}
Krishna Regmi and Mubarak Shah.
\newblock Bridging the domain gap for ground-to-aerial image matching.
\newblock In {\em The IEEE International Conference on Computer Vision (ICCV)},
  October 2019.

\bibitem{Riegler2020FVS}
Gernot Riegler and Vladlen Koltun.
\newblock Free view synthesis.
\newblock In {\em Eur. Conf. Comput. Vis.}, 2020.

\bibitem{ronneberger15unet}
Olaf Ronneberger, Philipp Fischer, and Thomas Brox.
\newblock U-net: Convolutional networks for biomedical image segmentation.
\newblock In {\em International Conference on Medical Image Computing and
  Computer-assisted Intervention}, pages 234--241. Springer, 2015.

\bibitem{saito2017temporal}
Masaki Saito, Eiichi Matsumoto, and Shunta Saito.
\newblock Temporal generative adversarial nets with singular value clipping.
\newblock In {\em Proceedings of the IEEE international conference on computer
  vision}, pages 2830--2839, 2017.

\bibitem{shih20203dp}
Meng-Li Shih, Shih-Yang Su, Johannes Kopf, and Jia-Bin Huang.
\newblock 3d photography using context-aware layered depth inpainting.
\newblock In {\em IEEE Conference on Computer Vision and Pattern Recognition
  (CVPR)}, 2020.

\bibitem{shum2000review}
Harry Shum and Sing~Bing Kang.
\newblock Review of image-based rendering techniques.
\newblock In {\em Visual Communications and Image Processing 2000}, volume
  4067, pages 2--13. International Society for Optics and Photonics, 2000.

\bibitem{vgg}
Karen Simonyan and Andrew Zisserman.
\newblock Very deep convolutional networks for large-scale image recognition.
\newblock In {\em International Conference on Learning Representations}, 2015.

\bibitem{sitzmann2019deepvoxels}
Vincent Sitzmann, Justus Thies, Felix Heide, Matthias Nie{\ss}ner, Gordon
  Wetzstein, and Michael Zollhofer.
\newblock Deepvoxels: Learning persistent 3d feature embeddings.
\newblock In {\em Proceedings of the IEEE Conference on Computer Vision and
  Pattern Recognition}, pages 2437--2446, 2019.

\bibitem{sitzmann2019scene}
Vincent Sitzmann, Michael Zollh{\"o}fer, and Gordon Wetzstein.
\newblock Scene representation networks: Continuous 3d-structure-aware neural
  scene representations.
\newblock In {\em Advances in Neural Information Processing Systems}, pages
  1121--1132, 2019.

\bibitem{tewari2020neuralrendering}
A. Tewari, O. Fried, J. Thies, V. Sitzmann, S. Lombardi, K. Sunkavalli, R.
  Martin-Brualla, T. Simon, J. Saragih, M. Nie{\ss}ner, R. Pandey, S. Fanello,
  G. Wetzstein, J.-Y. Zhu, C. Theobalt, M. Agrawala, E. Shechtman, D.~B
  Goldman, and M. Zollh{\"o}fer.
\newblock State of the art on neural rendering.
\newblock {\em Computer Graphics Forum (EG STAR 2020)}, 2020.

\bibitem{thies2019deferred}
Justus Thies, Michael Zollh{\"o}fer, and Matthias Nie{\ss}ner.
\newblock Deferred neural rendering: Image synthesis using neural textures.
\newblock {\em ACM Transactions on Graphics (TOG)}, 38(4):1--12, 2019.

\bibitem{tulyakov2018mocogan}
Sergey Tulyakov, Ming-Yu Liu, Xiaodong Yang, and Jan Kautz.
\newblock Mocogan: Decomposing motion and content for video generation.
\newblock In {\em Proceedings of the IEEE conference on computer vision and
  pattern recognition}, pages 1526--1535, 2018.

\bibitem{vondrick2016generating}
Carl Vondrick, Hamed Pirsiavash, and Antonio Torralba.
\newblock Generating videos with scene dynamics.
\newblock In {\em Advances in neural information processing systems}, pages
  613--621, 2016.

\bibitem{walker2016uncertain}
Jacob Walker, Carl Doersch, Abhinav Gupta, and Martial Hebert.
\newblock An uncertain future: Forecasting from static images using variational
  autoencoders.
\newblock In {\em European Conference on Computer Vision}, pages 835--851.
  Springer, 2016.

\bibitem{walker2017pose}
Jacob Walker, Kenneth Marino, Abhinav Gupta, and Martial Hebert.
\newblock The pose knows: Video forecasting by generating pose futures.
\newblock In {\em Proceedings of the IEEE international conference on computer
  vision}, pages 3332--3341, 2017.

\bibitem{wang2019few}
Ting-Chun Wang, Ming-Yu Liu, Andrew Tao, Guilin Liu, Bryan Catanzaro, and Jan
  Kautz.
\newblock Few-shot video-to-video synthesis.
\newblock {\em Advances in Neural Information Processing Systems},
  32:5013--5024, 2019.

\bibitem{wang2018vid2vid}
Ting-Chun Wang, Ming-Yu Liu, Jun-Yan Zhu, Guilin Liu, Andrew Tao, Jan Kautz,
  and Bryan Catanzaro.
\newblock Video-to-video synthesis.
\newblock In {\em Proceedings of Conference on Neural Information Processing
  Systems (NeurIPS)}, 2018.

\bibitem{wiles2020synsin}
Olivia Wiles, Georgia Gkioxari, Richard Szeliski, and Justin Johnson.
\newblock {SynSin}: {E}nd-to-end view synthesis from a single image.
\newblock In {\em CVPR}, 2020.

\bibitem{zhai2017predicting}
Menghua Zhai, Zachary Bessinger, Scott Workman, and Nathan Jacobs.
\newblock Predicting ground-level scene layout from aerial imagery.
\newblock In {\em Proceedings of the IEEE Conference on Computer Vision and
  Pattern Recognition}, pages 867--875, 2017.

\bibitem{zhu2017toward}
Jun-Yan Zhu, Richard Zhang, Deepak Pathak, Trevor Darrell, Alexei~A Efros,
  Oliver Wang, and Eli Shechtman.
\newblock Toward multimodal image-to-image translation.
\newblock In {\em Advances in Neural Information Processing Systems}, 2017.

\end{thebibliography}
}

\end{document}

%%%%%%%%% TITLE
\title{Sat2Vid: Street-view Panoramic Video Synthesis from a Single Satellite Image\\
Supplementary Materials}

\author{First Author\\
Institution1\\
Institution1 address\\
{\tt\small firstauthor@i1.org}
\and
Second Author\\
Institution2\\
First line of institution2 address\\
{\tt\small secondauthor@i2.org}
}

\maketitle
\ificcvfinal\thispagestyle{empty}\fi

\input{supp_tex/supp}

\clearpage

{
\small
\bibliographystyle{ieee_fullname}
\bibliography{sat2vid}
}